\documentclass{IEEEtran}
\usepackage[utf8]{inputenc}

\title{Immersive Operation of a Semi-Autonomous Aerial Platform for Detecting and Mapping Radiation}
\author{P. Dayani*, N. Orr*, A. Thomopoulos*, V. Saran, S. Krishnaswamy, E. Zhang, N. Hu, D. McPherson, J. Menke, A. Yang, K. Vetter 
\thanks{*These authors contributed equally.}\thanks{The project or effort depicted was or is sponsored by the Department of the Defense, Defense Threat Reduction Agency. The content of the information does not necessarily reflect the position or the policy of the federal government, and no official endorsement should be inferred. The project is also supported in part by a PCARI grant and an ONR grant N00014-19-1-2066.}\thanks{
P. Dayani, N. Orr, A. Thomopoulos, V. Saran, S. Krishnaswamy, E. Zhang, N. Hu, D. McPherson, J. Menke, and A. Yang are with the EECS Department, 
K. Vetter is with the Nuclear Engineering Department at the University of California, Berkeley, Berkeley, CA 94720 USA.}}
\date{May 2020}

\usepackage{graphicx}

\begin{document}

\maketitle

\begin{abstract}
    Recent advancements in radiation detection and computer vision have enabled small unmanned aerial systems (sUASs) to produce 3D nuclear radiation maps in real-time. Currently these state-of-the-art systems still require two operators: one to pilot the sUAS and another operator to monitor the detected radiation. In this work we present a system that integrates real-time 3D radiation visualization with semi-autonomous sUAS control. Our Virtual Reality interface enables a single operator to define trajectories using waypoints to abstract complex flight control and utilize the semi-autonomous maneuvering capabilities of the sUAS. The interface also displays a fused radiation visualization and environment map, thereby enabling simultaneous remote operation and radiation monitoring by a single operator. This serves as the basis for development of a single system that deploys and autonomously controls fleets of sUASs.

\end{abstract}

\section{Introduction}

The ability to accurately detect and monitor radiological materials remains critical in nuclear security and safeguards, in emergency response, and in the operation, decommissioning, and remediation of nuclear facilities. These applications require an effective visualization of radiation data to allow operators to understand the situation and make informed decisions. 

Radiation can now be effectively mapped by implementing 3D Scene Data Fusion (SDF) \cite{1a} to combine 3D radiation and environment data. The Localization and Mapping Platform (LAMP), a lightweight device that executes SDF, can be deployed onboard an sUAS \cite{1c}. A LAMP (referring to any of its variants such as the Neutron Gamma LAMP or the LaBr LAMP) has performed SDF onboard an sUAS, leading to unprecedented capabilities \cite{1b} in the search for and response to the release of radiological material. An sUAS can efficiently map areas with challenging terrain or dangerous conditions. In addition, flying near hard-to-reach surfaces generates detailed volumetric radiation mapping without endangering the operator. 

Similar real-time systems require two independent operators-- one controlling the sUAS and another interpreting radiation readings on a 2D monitor \cite{1c}. These two-operator systems are limited by a lack of stereoscopic depth cues on traditional 2D monitors when viewing 3D data. The requirement for the two operators to manually synchronize mission-critical data may also compromise real-time response to time-critical information.

\begin{figure}[t!]
\centering
\includegraphics[width=\columnwidth]{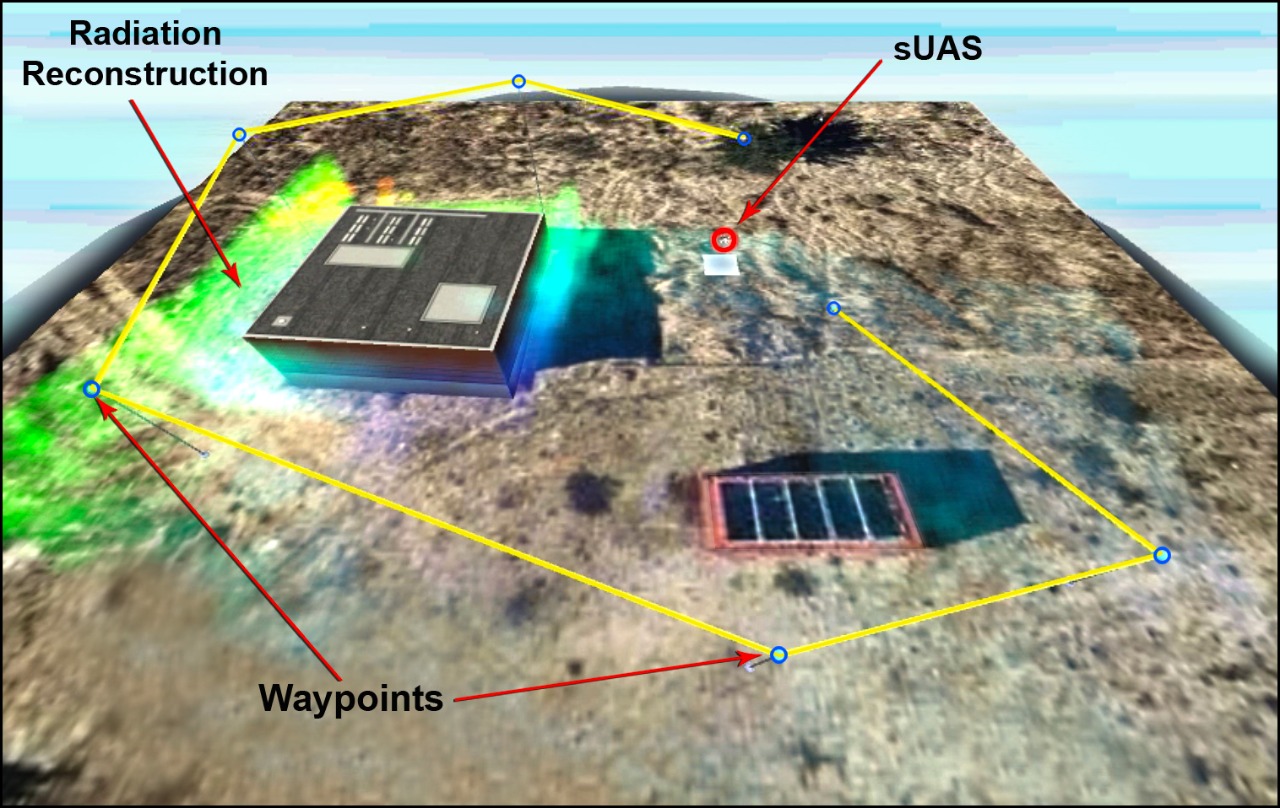}
\caption{Our VR interface displays a textured 3D satellite view of the environment. The operator sets waypoints (blue targets) through the VR interface to mark important geolocations. The sUAS (red) follows the flight path (yellow) around the building, detecting and streaming radiation and environment reconstruction data back to the interface as a colorized 3D mesh.}
\label{fig:keyfigure}
\end{figure}

Virtual Reality (VR) overcomes these limitations. Stereoscopic depth cues intrinsic to VR convey enhanced spatial understanding of 3D data, which is critical for remote sUAS control \cite{4e, 4a, 3c}. Furthermore, remote sUAS control in VR offers a more intuitive and efficient flight planning experience \cite{2b} for operators without in-depth knowledge of sUAS systems \cite{2c}. Integrating data visualization with sUAS control in a VR interface enables the operator to make informed decisions without relying on a separate pilot or interpreter.

In this work we present a system that integrates real-time 3D radiation mapping with semi-autonomous sUAS control. We leverage widely available commercial VR and drone navigation systems to enable a sole operator to safely conduct radiation detection missions as shown in Figure \ref{fig:keyfigure}. 

\section{System}

\subsection{Hardware and Software Overview}

\begin{figure}[h!]
\centering
\includegraphics[scale=0.4]{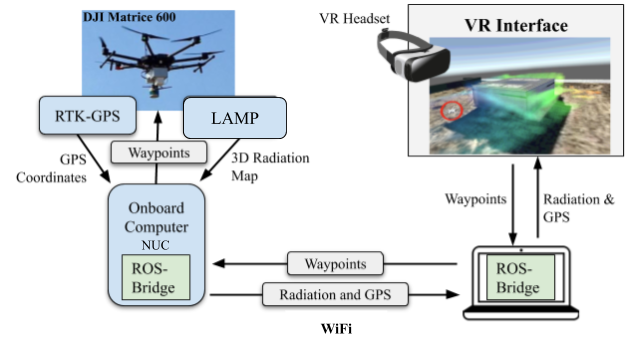}
\caption{Workflow of deploying a radiation-mapping mission between an sUAS (left) and the VR interface (right). Waypoints are placed on the 3D map in VR and transmitted over WiFi as geolocated flight targets for the sUAS to reach semi-autonomously. Operators can view the real-time LAMP radiation and surface reconstruction from the sUASs in VR.}
\label{fig:systemdiagram}
\end{figure}

Our system consists of a semi-autonomous radiation detecting sUAS that communicates with a remote VR interface as shown in Figure \ref{fig:systemdiagram}. The VR interface allows the operator to analyze mapped radiation and control the sUAS.

We chose the DJI Matrice 600 as our sUAS for its semi-autonomous capabilities and extensible 6\,kg payload capacity. The sUAS carries a Real-Time Kinematic GPS (RTK-GPS) and a LAMP with an onboard Intel NUC computer. During a mission, the RTK-GPS localizes the sUAS to within a few centimeters, enabling granular control. Meanwhile, the LAMP performs SDF to integrate and produce an accurate 3D representation of surrounding radiation and environmental data. The onboard computer establishes an interface between the LAMP, RTK-GPS, and the VR application, as shown in Figure \ref{fig:systemdiagram}. A two-way communication channel with the VR application is established using the Robot Operating System (ROS) Bridge Library over WiFi. Geolocated waypoints received from the VR application are used by the Matrice 600’s navigation system for semi-autonomous flight while the LAMP radiation data and GPS data are constantly transmitted over this channel back to the VR interface. 

The VR interface is built on Unity and deployed on any Oculus headset. The operator accesses the immersive experience through the headset and can interact with the virtual environment using two hand-held VR controllers.

\subsection{Virtual Reality User Interface}

The key advantage of our platform is the combination of 3D radiation visualization and sUAS control in a single integrated virtual reality interface.

As shown in Figure \ref{fig:keyfigure}, the VR environment is initially composed of a 3D satellite view of the sUAS's environment. The operator can then orient themselves within the virtual environment by panning, rotating, and zooming the map. Assistive visuals such as battery levels, GPS signal strength, and other interface elements can be toggled on or off. The visualized radiation overlay, environment, satellite imagery, and sUAS all scale together to always provide an accurate representation of the real world to the operator.

The operator can command the sUAS to traverse specific geo-locations in the 3D environment. As the sUAS flies in the real world, radiation from the environment is visualized in the interface as a colorized 3D mesh depicting varying radiation intensity levels (Figure \ref{fig:keyfigure}). This real-time feedback enhances the operator’s ability to precisely assess the current situation and manage the mission accordingly.

\subsection{Waypoint Placement System}

Our waypoint system allows the operator to control the semi-autonomous sUAS. In the virtual 3D map the operator marks a sequence of locations, each denoting a waypoint, and the sUAS flight trajectory is automatically interpolated in between as shown in Figure \ref{fig:keyfigure}. 

The operator can also edit each waypoint as needed, and then they can initiate the sUAS mission when satisfied with the flight path. The VR interface converts the in-app waypoint locations to real-world GPS coordinates. Subsequently, the sUAS executes the flight path, flying to each waypoint in the set order. During sUAS flight, the operator can still edit any future waypoints to update the flight path upon analyzing the live-streamed radiation.

With waypoint missions, the operator only needs to select important targets for the sUAS. Our system abstracts away complex controls such as flight stabilization, reducing the operator's cognitive load and need for piloting expertise. 

\section{Conclusion}
We introduce a novel system that performs simultaneous radiation detection, visualization, and sUAS flight control. Existing systems have separate mapping and control systems, therefore relying on two operators. Our system enhances a single operator's spatial awareness by displaying 3D radiation data in the same interface as the controls, thus eliminating the need for a second operator. 

In the future, we will enhance the virtual reality interface, real-world sUAS flight, data collection, and visualization procedures of our system. We believe that our system will help make radiation mapping missions safer, easier, and more efficient. Our system can be extended to control multiple sUASs for radiation mapping  in large geographical areas that are inaccessible or dangerous for humans to explore. Furthermore, our work can also be expanded to enable long-range, beyond visual line of sight (BVLOS) flight missions. BVLOS missions can leverage autonomous obstacle detection for increased sUAS safety. Adding a textured 3D real-time map of the environment to the VR interface will also improve safety by increasing the operator's understanding of the sUAS's surroundings. Finally, our system can be expanded to include automatic waypoint-placement algorithms for more efficient radiation detection and source localization.

\bibliographystyle{IEEEtran}
\bibliography{references}
\end{document}